\newcommand{\CLASSINPUTinnersidemargin}{1in}
\newcommand{\CLASSINPUToutersidemargin}{1in}
\newcommand{\CLASSINPUTtoptextmargin}{1in}
\newcommand{\CLASSINPUTbottomtextmargin}{1in}

\documentclass[11pt,letterpaper,conference,final]{IEEEtran}
\IEEEoverridecommandlockouts
\usepackage{url} 
\usepackage[hidelinks]{hyperref}

\usepackage{cite}
\usepackage{amsmath,amssymb,amsfonts}
\usepackage{algorithmic}
\usepackage{graphicx}
\usepackage{textcomp}
\usepackage{xcolor}
\def\BibTeX{{\rm B\kern-.05em{\sc i\kern-.025em b}\kern-.08em
    T\kern-.1667em\lower.7ex\hbox{E}\kern-.125emX}}

\begin{document}

\title{Mars Spectrometry 2: Gas Chromatography - Second place solution
}

\author{\IEEEauthorblockN{Dmitry A. Konovalov}
\IEEEauthorblockA{\textit{College of Science and Engineering}, 
\textit{James Cook University}\\
Townsville, Australia, 
dmitry.konovalov@jcu.edu.au}
}
\maketitle
\thispagestyle{plain}
\pagestyle{plain}

\begin{abstract}

The Mars Spectrometry 2: Gas Chromatography challenge was sponsored by NASA and run on the DrivenData competition platform in 2022. This report describes the solution which 
achieved the second-best score on the competition's test dataset. 
The solution utilized two-dimensional, image-like representations of  
the competition's chromatography data samples. 
A number of different 
Convolutional Neural Network models were trained and ensembled for the final submission.

\end{abstract}

% \begin{IEEEkeywords}
% component, formatting, style, styling, insert
% \end{IEEEkeywords}

% \section{Performance}

\section{Challenge summary}
The goal of the {\em Mars Spectrometry 2: Gas Chromatography} challenge \cite{Mars2} was to develop a model, 
which could automatically process supplied gas chromatography-mass spectrometry (GCMS) data files.
The challenge was set up as a supervised multi-label classification
machine learning problem, where for each {\em training} sample data file, nine binary nonexclusive
{\em target} labels were provided. The challenge participants were required to predict the target labels for a test dataset for which the ground-truth labels were not supplied. 
Submitted test predictions were evaluated by multilabel aggregated {\em log loss} score (a lower value is better), which penalized confident but incorrect predictions. 

The first place solution achieved $0.1443$, this second place solution had $0.1485$ ({\bf +2.9}\% higher relative to the first place), closely followed by the third $0.1497$ ({\bf +3.7\%}). 
The corresponding three best scores from 
the first Mars Spectrometry (Mars-1) competition \cite{Mars1} were:
$0.0920$ (1st place), $0.1160$ ({\bf +26\% }) and $0.1189$ ({\bf +29\%}), 
where the same {\em log loss} metric was used.

Two observations could be made by comparing the Mars-1 and Mars-2 final leaderboards. First, the top three solutions of Mars-2 are comparable within 4\% of the used performance metric, while the first-place solution of Mars-1 had a 26\% improvement gap compared to the second-best solution. 
This superiority gap motivated me to focus on applying the first-place solution of Mars-1 
\cite{Mars1st} to the Mars-2 challenge.

The second observation was due to the 56\% deterioration of the winning Mars-2 score ($0.1443$) 
compared to the corresponding Mars-1 score ($0.0920$), indicating that Mars-2 was a "harder" challenge to solve.

\section{Solution development}
In Mars-2, each training and test data file was a CSV file containing three columns: {\em time} (in minutes), {\em mass} (mass-to-charge ration, {\em m/z}) 
and {\em intensity} of detected ions per second in arbitrary but relative (within one 
sample) units. The main critical difference of Mars-2 samples (compared to Mars-1) 
was the absence of temperature values. As per the competition's problem description, 
time values did not have any spectrometry significance by themselves
and were only proxies for missing temperature values. 
Furthermore, the only thing we could assume was that sample temperature increased with time, 
{\em "but the temperature ramp is not exactly known nor the same across samples"} \cite{Mars2}.

This solution evolved from the published first-place solution of the predecessor 
Mars-1 competition \cite{Mars1},
{\em Mars Spectrometry: Detect Evidence for Past Habitability} \cite{Mars1st}. 
The CSV sample files from the training dataset 
were converted to 2D images and used for training (or fine-tuning) 
Deep Learning models such as Convolutional Neural Networks (CNNs). 
The ImageNet \cite{deng2009imagenet} pre-trained, PyTorch-based \cite{NEURIPS2019_9015} backbones and models were used from the timm \cite{timm} python package. 

Working within the limits of not knowing the spectrometry-relevant temperatures, 
time values were scaled in all samples to the [0,1] range and then binned to a configurable number of time slots $N_{t}$, where $N_{t}=192$ was used in the final submission. 
When training a CNN, the time dimension was randomly batch-wise resized on GPU within the 128-256 range of values. 
Then, at the inference phase, the corresponding TTA (test-time-augmentation) was done by averaging time sizes (5 steps of 32, centred at 192). 
Such time resizing during training and testing (via TTA) assisted in capturing some 
temperature-related data features. 
However, samples with "slower" ramping up to "lower" (than the training mean) 
temperatures as well as  samples with "faster" ramping up to "higher" temperatures 
would be unlikely to yield correct predictions within this solution. 
It is possible that the large variations in the ramping rates together 
with the large variations in the actual final temperatures were the main reason 
why Mars-2 leaderboard scores were much worse than the corresponding Mars-1 values. Hence:

{\em \textbf{Actionable recommendation:} 
Even if the exact time-temperature ramp functions are not known, 
the availability of start and end temperature values (per sample) 
should greatly improve this solution.}

For example, if the maximum temperature of one sample $T_i$ is double the 
temperature of another sample $T_j$, $T_i = 2 * T_j$, then the $i$'th sample 
time values should be binned into a proportionally larger number of time slots.
Furthermore, the variations in the ramping functions could be modelled by 
random time warping, for example via 
\begin{equation}
f(t) = t^{\alpha}, \ \ \ 0.5 \le \alpha \le 2,  \ \ \ 0 \le t \le 1, 
\end{equation}
which was also attempted in this solution but was not fully explored to verify its 
utility. 

\begin{figure}[htbp]
\centering
\includegraphics[width=0.45\textwidth]{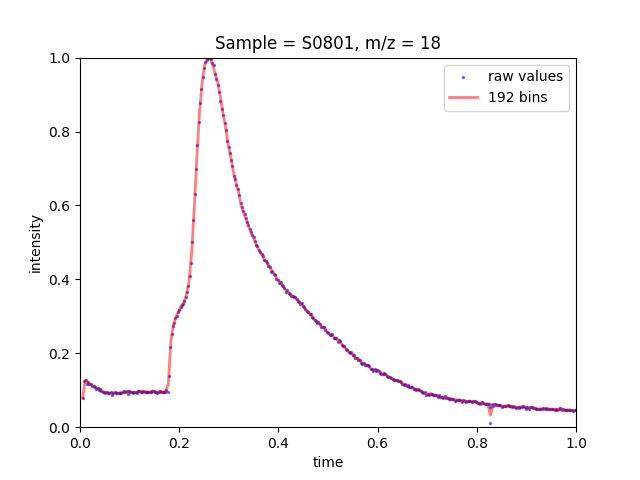}
\caption{Raw intensity values for sample S0801 at m/z=18}
\label{fig1}
\end{figure}

\begin{figure}[htbp]
\centering
\includegraphics[width=0.25\textwidth]{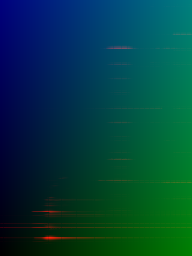}
\caption{Sample S0801 converted to 2D representation and saved into red channel, 
where y-axis is mass rows ($0 \le m \le 255$) and x-axis is time columns ($0 \le t \le 191$). The green and blue channels are loaded with $t/191$ and $m/255$ as per \cite{Mars1st}, correspondingly. }
\label{fig2}
\end{figure}

\begin{figure}[htbp]
\centering
\includegraphics[width=0.25\textwidth]{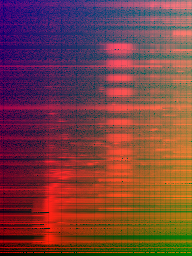}
\caption{Same as in Fig.~\ref{fig2} but divided by maximum column values (mass-normalization).}
\label{fig3}
\end{figure}

\begin{figure}[htbp]
\centering
\includegraphics[width=0.25\textwidth]{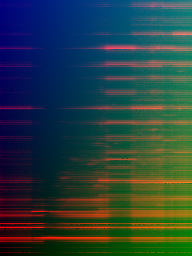}
\caption{Same as in Fig.~\ref{fig2} but divided by maximum row values (time-normalization).}
\label{fig4}
\end{figure}

\begin{figure}[htbp]
\centering
\includegraphics[width=0.4\textwidth]{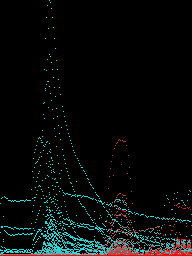}
\caption{Same as in Fig.~\ref{fig2} but with swapped mass and intensity values, and without the mass and time positional channel encodings. 
The red color component was proportional to mass, where a mass value of 255 corresponded to the 
maximum color red value of 255 (in uint8 RGB color encoding).}
\label{fig5}
\end{figure}

\subsection{Two-dimensional representations}

Dot-scatter plot in Fig.~\ref{fig1} displays the raw intensity values for $m/z=18$ in one of the samples. The red line is the same values averaged into 192 equally spaced time slots. 
Note that the raw values and the 192-bin smoothed values are very similar.

Fig.~\ref{fig2} displays the actual 2D representation of the same sample for all mass and time values, where 256 mass bins (image rows) and 192 time slots (image columns) were used.
Mass (i.e. $m/z$) values were rounded to the nearest integer values retaining only the first 256 values. The positional encoding of mass and time dimensions into separate image channels was adopted from the Mars-1 first-place solution \cite{Mars1st}.

To reveal feature-rich patterns in the 2D representations, the images could be 
divided by maximum column (mass-normalization \cite{Mars1st}, Fig.~\ref{fig3}) 
or row (time-normalization Fig.~\ref{fig4}) values. 
In fact, all the models used in this solution's final submission 
were either mass or time normalized.

All intensity values were converted to a 
logarithmic scale via either one of the following two ways
\begin{equation}
\log_{10}(1 + 10^3 \times a), \ \ 0 \le a \le 1, 
\end{equation}
\begin{equation}
\log_{10}(a), \ \ 10^{-4} \le a \le 1. 
\end{equation}

Another completely different (from the ones considered above, see Figs.~\ref{fig2}-\ref{fig4}) conversion 
to a 2D representation (see Fig.~\ref{fig5}) 
was attempted but did not contribute to 
the final ensembling. In that conversion variation, mass values were encoded as the 
pixel color values while the intensity values were binned into 256 y-axis values (image rows).
It is likely that this approach could be made useful with further development
as it represents the data samples from a different point of view.

\subsection{Models and training pipeline}

I followed the first-place solution \cite{Mars1st} 
of the previous Mars challenge \cite{Mars1} 
by converting the mass spectrometry data into 2D images. 
A total of 13 CNN models and data-processing configurations were ensembled for the final submission. For ensembling, averaging logits (inverted $10^{-4}$ clipped sigmoids) rather than probabilities improved both validation and test leaderboard results. 

I experimented with different conversion configurations (see preceding subsection) and found that the first 256 mass values (y-axis) were sufficient. For the time axis (x-axis), 192 time slots were selected as a reasonable baseline value, where larger values slowed down the CNN training.

In this competition, time was only a proxy for temperature. Therefore, I only explored ideas where exact dependence on time values was not required. That led to two key ideas, which pushed this solution to the winning range. First was, as explained earlier, the random resizing of the 
time dimension with the corresponding TTA. The second win-contributing idea
was creating a custom time-averaged head, 
where only the time dimension of the CNN backbone 2D features was averaged 
(rather than both the mass and time dimensions) before the last fully connected liner layer. 

For the second (standard model) architecture, a small improvement was gained by using the full pre-trained timm models before the required 9-label linear classifier, where 
the ImageNet classification 
heads were retained rather than just the models' backbones.

An extensive search of timm \cite{timm} pre-trained models found HRNet-w64 
\cite{HRNET} to be particularly accurate for 
the considered 2D representations of data. 
The following other timm's pre-trained backbones were also used: dpn98 and dpn107 \cite{DPN}, regnetx\_320 \cite{RegNetX} and resnet34. 

Small but consistent improvement was gained by encoding the {\em derivatized} meta-data column as a 2-channel image and adding a trainable conversion layer before a CNN backbone.

While the original Mars-1 solution \cite{Mars1st} used noise 
augmentations extensively,
I was not able to achieve any consistent validation/OOF
loss improvement by 
adding noise and/or smoothing/pre-processing original data at the training and/or inference stages. Hence, all training and inference were done with the original data converted to 2D images (as described in the preceding subsection) 
without any further processing.

All models were trained for 20 epochs with a cosine learning rate schedule and linear 2-epoch warm-up and a base learning rate of $10^{-4}$. About half of the ensembled models were trained with mixup \cite{zhang2017mixup} (a probability of 0.1).

\begin{figure}[htbp]
\centering
\includegraphics[width=0.24\textwidth]{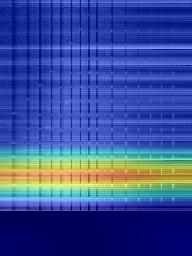}
\includegraphics[width=0.24\textwidth]{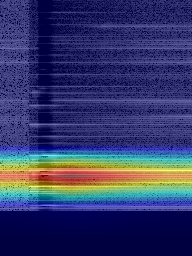}
\caption{Grad-CAM++ visualisation of two samples (S0047 and S0078) 
containing {\em hydrocarbon} compounds.}
\label{fig_interp}
\end{figure}

\begin{figure}[htbp]
\centering
\includegraphics[width=0.24\textwidth]{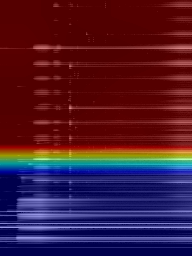}
\includegraphics[width=0.24\textwidth]{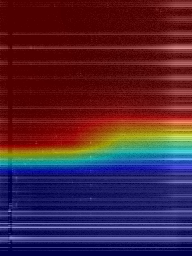}
\caption{Same as Fig.~\ref{fig_interp} but for S0107 and S0305 samples 
containing {\em mineral} compounds.}
\label{fig_interp2}
\end{figure}

\section{Interpretability/Explainabilty}

Bonus algorithm explainability awards were run as part of DrivenData's "Where's Whale-do?" competition \cite{DDWhales}. 
One of the winners of that bonus round was the 4th place submission \cite{Whale4th}, 
which utilized the Grad-CAM \cite{jacobgilpytorchcam} package containing the state-of-the-art methods for {\em Explainable AI for computer vision}.
The Grad-CAM approach of \cite{Whale4th} was replicated here to illustrate its 
feasibility to explain the solution's predictions.

The best out-of-fold (OOF) model of this solution was the custom time-averaged head 
on top of the head-less HRNet-w64 \cite{HRNET} backbone, 
where the ImageNet pre-trained HRNet-w64 backbone was available in the timm \cite{timm} package. The Grad-CAM++ \cite{chattopadhay2018grad} implementation
from \cite{jacobgilpytorchcam} was used to calculate the gradients of the backbone's
last 2D layer. 
Fig.~\ref{fig_interp} shows two OOF samples 
ground-truth labelled to contain only 
{\em hydrocarbon} compounds, where they were correctly and confidently (probability $> 0.9$) predicted by the model. 
In all figures except Fig.~\ref{fig5}, 
$m/z=0$ is the bottom row of pixels and $m/z=255$ is the top row.
Comparison of Fig.~\ref{fig_interp} and Fig.~\ref{fig_interp2}, 
revealed that the mineral compounds (Fig.~\ref{fig_interp2}) 
were classified by the model due to the ions with $m/z>100$, 
while the ions around $m/z=80$ were the main reason for predicting
the hydrocarbon compounds. 
Also, note that the mineral compounds had nearly uniform activations (red-hue colors) 
across the time dimension (x-axis), while hydrocarbons were more activated at 
lower temperatures (smaller time values).

% \url{https://www.drivendata.org/competitions/97/nasa-mars-gcms} is 

% \section{Interpretability}

% \section{Innovation}

% \section{Constructive discussion}

% \begin{table}[htbp]
% \caption{Table Type Styles}
% \begin{center}
% \begin{tabular}{|c|c|c|c|}
% \hline
% \textbf{Table}&\multicolumn{3}{|c|}{\textbf{Table Column Head}} \\
% \cline{2-4} 
% \textbf{Head} & \textbf{\textit{Table column subhead}}& \textbf{\textit{Subhead}}& \textbf{\textit{Subhead}} \\
% \hline
% copy& More table copy$^{\mathrm{a}}$& &  \\
% \hline
% \multicolumn{4}{l}{$^{\mathrm{a}}$Sample of a Table footnote.}
% \end{tabular}
% \label{tab1}
% \end{center}
% \end{table}

% \section*{Acknowledgment}
% The preferred spelling of the word ``acknowledgment'' in America is without 
% an ``e'' after the ``g''. Avoid the stilted expression ``one of us (R. B. 
% G.) thanks $\ldots$''. Instead, try ``R. B. G. thanks$\ldots$''. Put sponsor 
% acknowledgments in the unnumbered footnote on the first page.

% \section*{References}
% \bibliographystyle{plain}
% \bibliography{refs.bib}

\bibliographystyle{IEEEtran}
\bibliography{IEEEabrv,refs}

\end{document}